\pdfoutput=1

\documentclass[11pt]{article}

\usepackage[final]{acl}

\usepackage{times}
\usepackage{latexsym}
\usepackage{booktabs}
\usepackage{amsmath}
\usepackage{bm}
\usepackage{amssymb}
\usepackage{mathtools}
\usepackage{amsthm}
\usepackage{fontawesome}
\usepackage{subcaption}

\newcommand{\vv}{\mathbf{v}}
\newcommand{\vk}{\mathbf{k}}
\newcommand{\vh}{\mathbf{h}}
\newcommand{\vo}{\mathbf{o}}
\newcommand{\vr}{\mathbf{r}}
\newcommand{\vd}{\mathbf{d}}
\newcommand{\vq}{\mathbf{q}}
\newcommand{\vi}{\mathbf{i}}
\newcommand{\vx}{\mathbf{x}}
\newcommand{\balpha}{\bm{\alpha}}

\usepackage[T1]{fontenc}

\usepackage[utf8]{inputenc}

\usepackage{microtype}

\usepackage{inconsolata}

\usepackage{graphicx}

\title{A Systematic Analysis of Hybrid Linear Attention}

\author{
 \textbf{Dustin Wang*\textsuperscript{1}},
 \textbf{Rui-Jie Zhu*\textsuperscript{1,2}},
 \textbf{Steven Abreu\textsuperscript{3}},
 \textbf{Yong Shan\textsuperscript{2}},
 \textbf{Taylor Kergan\textsuperscript{1}},
\\
 \textbf{Yuqi Pan\textsuperscript{2,4}},
 \textbf{Yuhong Chou\textsuperscript{5}},
 \textbf{Zheng Li \textsuperscript{2}},
 \textbf{Jibin Wu\textsuperscript{5}},
\\
 \textbf{Ge Zhang\textsuperscript{2, 6, \textdagger}},
 \textbf{Wenhao Huang\textsuperscript{2}},
 \textbf{Jason Eshraghian\textsuperscript{1, \textdagger}},
\\
\\
 \textsuperscript{1}UC Santa Cruz,
 \textsuperscript{2}ByteDance Seed,
 \textsuperscript{3}University of Groningen,
 \textsuperscript{4}CASIA,
 \textsuperscript{5}PolyU,
 \textsuperscript{6}M-A-P
\\
\\
 *Equal Contribution,
 \textsuperscript{\textdagger}Corresponding Authors
\\
}

\begin{document}
\maketitle
\begin{abstract}
Transformers face quadratic complexity and memory issues with long sequences, prompting the adoption of linear attention mechanisms using fixed-size hidden states. However, linear models often suffer from limited recall performance, leading to hybrid architectures that combine linear and full attention layers. Despite extensive hybrid architecture research, the choice of linear attention component has not been deeply explored. We systematically evaluate various linear attention models across generations—vector recurrences to advanced gating mechanisms with both standalone and hybridized. To enable this comprehensive analysis, we trained and will open-source 72 models\footnote{ Model training and inference are open-sourced at https://huggingface.co/collections/m-a-p/hybrid-linear-attention-research.}: 36 at 340M parameters (20B tokens) and 36 at 1.3B parameters (100B tokens), covering six linear attention variants across five hybridization ratios. Benchmarking on standard language modeling and recall tasks reveals that superior standalone linear models do not necessarily excel in hybrids. While language modeling remains stable across linear-to-full attention ratios, recall significantly improves with increased full attention layers, particularly below a 3:1 ratio. Our study highlights selective gating, hierarchical recurrence, and controlled forgetting as critical for effective hybrid models. We recommend architectures such as HGRN-2 or GatedDeltaNet with a linear-to-full ratio between 3:1 and 6:1 to achieve Transformer-level recall efficiently.
\end{abstract}

\section{Introduction}

The Transformer architecture \cite{vaswani2023attentionneed} has become the dominant network design for Large Language Models (LLMs). However, its quadratic complexity in sequence length for a single forward pass, $O(L^2)$, and the linearly growing Key-Value (KV) cache in full attention mechanisms \cite{vaswani2023attentionneed}, become problematic as sequences lengthen. To address this, numerous alternative approaches have emerged, with linear complexity models representing one of the most promising directions \cite{katharopoulos2020transformersrnnsfastautoregressive}. These models compress the linearly growing KV cache into a static vector or matrix—a hidden state—thereby achieving $O(L)$ overall complexity \cite{katharopoulos2020transformersrnnsfastautoregressive}. This innovation tackles both the quadratic complexity problem and the issue of an expanding KV cache.

To better utilize these fixed hidden states, linear complexity models have evolved through roughly three generations. The first generation was characterized by vector-level hidden states derived from traditional Recurrent Neural Networks (RNNs)~\cite{voelker2019legendre}. The second generation advanced this by employing outer products for state expansion, creating 2D hidden states \cite{qin2024hgrn2gatedlinearrnns}. The third generation further refined these models by introducing delta rule state transitions \cite{yang2025parallelizinglineartransformersdelta}, replacing the hidden state decay mechanisms used in the first two generations. Each successive generation has generally yielded improved performance in both common reasoning and retrieval benchmarks.

Despite these advancements, the same fixed hidden state that enables linear complexity also introduces limitations in recall capabilities; linear attention models typically demonstrate poorer recall performance~\citep{shen2024scaling,jelassi2024repeat,sieber2024understanding}. To mitigate this deficiency, Hybrid Linear Attention models have emerged~\citep{team2024jamba}. This approach interleaves full attention layers with linear attention layers. Although the overall complexity remains quadratic due to the inclusion of full attention, these hybrid models deliver significantly better recall performance~\citep{dong2024hymba,minimax2025minimax01scalingfoundationmodels} and have successfully scaled to 400 billion parameters while maintaining competitive common reasoning performance~\citep{minimax2025minimax01scalingfoundationmodels}.

However, a critical observation in the development of hybrid models is that while the choice of the full attention mechanism is generally consistent, the selection of the linear attention component appears relatively arbitrary. Existing research predominantly focuses on ablation studies concerning the ratio of full to linear attention layers, rather than on the specific architectural choices within the linear models themselves \cite{lieber2024jamba, dong2024hymba, minimax2025minimax01scalingfoundationmodels, stripedhyena}. These studies often implicitly assume that a superior standalone linear model will inherently lead to a better-performing hybrid model. We ask: is this assumption valid? Furthermore, the determination of the optimal ratio in hybrid models is frequently guided by minimizing loss values, which primarily reflects performance on short-text modeling tasks, potentially overlooking the more critical aspect of recall performance. Finally, Ref.~\cite{liu2025scaling} has demonstrated that simply increasing the state size of linear attention models yields diminishing returns. Given this limitation, we hypothesize that performance improvements may instead depend on a model's ability to effectively manage and retrieve information from its stored memory--a capability we examine through our language modeling and recall evaluations.

In this work, we conduct a comprehensive investigation into a wide array of linear attention architectures spanning all three generations, analyzing their language modeling and recall performance when hybridized. We also benchmark pure linear attention models as a baseline and examine the impact of varying the ratio between full and linear attention layers. Our research aims to answer three primary questions:
\textbf{1)} Does a linear attention model with better standalone performance necessarily translate to a superior hybrid model architecture?
\textbf{2)} What aspects are more significantly influenced by the ratio of full to linear attention layers: overall language modeling (LM) performance or recall capabilities?
\textbf{3)} Which architectural components or design principles within linear attention models are most crucial for optimizing both LM performance and recall?

This paper makes three primary contributions. 
\begin{enumerate}
    \item We provide a comparative analysis of various linear attention architectures from all three recognized generations, evaluating both their standalone language modeling and their recall performance and also when integrated into hybrid models;
    \item We examine the assumption that improvements in standalone linear attention models directly correlate with enhanced performance in hybrid architectures, offering evidence-based insights into this relationship;
    \item Our work analyzes how the ratio of full attention to linear attention layers in hybrid models distinctly affects overall language modeling  capabilities versus specific recall performance, while also identifying key architectural components within linear attention models that are crucial for these metrics.
\end{enumerate}

\section{Related work}

\label{sec:related}
\renewcommand{\arraystretch}{1.25} 
\begin{table*}[t]
\centering
\footnotesize
\begin{tabular}{lll}
\toprule
\textbf{Model} & \textbf{Update rule ($\mathrm{S}_t$)} & \textbf{Read-out ($\vo_t$)} \\
\midrule
\multicolumn{3}{c}{\textbf{Vector-valued hidden state (classical / gated RNNs)}}\\
\midrule
HGRN   
            & $\vh_t = \balpha_t \odot \vh_{t-1} + (1-\balpha_t)\odot\vv_t$
            & $\vo_t = \vh_t \odot \vq_t$ \\
Hawk (RG-LRU)
            & $\vh_t = \vr_t\,\vh_{t-1} + \vi_t \odot \vx_t$
            & $\vo_t = \vh_t \odot \vq_t$ \\
\midrule
\multicolumn{3}{c}{\textbf{Matrix-valued state via outer products}}\\
\midrule
RetNet/Lightning     
            & $\mathrm{S}_t = \gamma\,\mathrm{S}_{t-1} + \vv_t\vk_t^\top$
            & $\vo_t = \mathrm{S}_t \vq_t$ \\

GLA  
            & $\mathrm{S}_t = \mathrm{S}_{t-1}\odot(\mathbf1\balpha_t^\top) + \vv_t\vk_t^\top$
            & $\vo_t = \mathrm{S}_t \vq_t$ \\
Mamba-2     
            & $\mathrm{S}_t = \gamma_t\,\mathrm{S}_{t-1} + \vv_t\vk_t^\top$
            & $\vo_t = \mathrm{S}_t \vq_t$ \\

RWKV-6   
            & $\mathrm{S}_t = \mathrm{S}_{t-1}\mathrm{Diag}(\balpha_t) + \vv_t\vk_t^\top$
            & $\vo_t = (\mathrm{S}_{t-1}+(\vd\odot\vv_t)\vk_t^\top)\vq_t$ \\
HGRN-2/MetaLA     
            & $\mathrm{S}_t = \mathrm{S}_{t-1}\mathrm{Diag}(\balpha_t) + \vv_t(\mathbf1-\balpha_t)^\top$
            & $\vo_t = \mathrm{S}_t \vq_t$ \\
\midrule
\multicolumn{3}{c}{\textbf{Delta-rule / controlled-forgetting family}}\\
\midrule
DeltaNet   
            & $\mathrm{S}_t = \mathrm{S}_{t-1}(\mathbf I-\beta_t\vk_t\vk_t^\top)+\beta_t\vv_t\vk_t^\top$
            & $\vo_t = \mathrm{S}_t \vq_t$ \\
Gated DeltaNet 
            & $\mathrm{S}_t = \alpha_t\,\mathrm{S}_{t-1}(\mathbf I-\beta_t\vk_t\vk_t^\top)+\beta_t\vv_t\vk_t^\top$
            & $\vo_t = \mathrm{S}_t \vq_t$ \\
\bottomrule
\end{tabular}

\caption{Unified comparison of linear-time attention mechanisms across three “generations”.
1st Gen keeps a vector $\vh_t\!\in\!\mathbb R^{d}$ with element-wise gating e.g.\ HGRN \cite{qin2024hgrn2gatedlinearrnns} and Hawk \cite{de2024griffin}, so the read-out is a Hadamard product ($\vh_t\odot\vq_t$).
2nd Gen stores a matrix $\mathrm{S}_t\!\in\!\mathbb R^{d\times n}$ via outer-product additions plus multiplicative decay, including GLA \cite{yang_gated_linear_attention}, RetNet \cite{sun2023retentivenetworksuccessortransformer,minimax2025minimax01scalingfoundationmodels}, RWKV-6 \cite{peng2024eaglefinchrwkvmatrixvalued}, HGRN-2 \cite{qin2024hgrn2gatedlinearrnns,chou2024metala} and Mamba \cite{dao2024mamba2}.
3rd Gen introduces delta-rule controlled forgetting that explicitly erases stale content \cite{yang2025parallelizinglineartransformersdelta, yang2025gateddeltanetworksimproving}.
$\vv_t,\vk_t,\vq_t$ are value, key, and query projections; $\balpha_t,\beta_t,\vr_t,\vi_t$ are gates; $\odot$ is the Hadamard product.}
\label{tab:gen-comparison}
\end{table*}

\paragraph{Linear complexity language models.}  
Early efforts to tame the quadratic cost of softmax attention rewrote it as an associative recurrent update, giving linear attention with $O(L)$ compute and memory \citep{katharopoulos2020transformersrnnsfastautoregressive}.  Subsequent fast-weight and kernel views improved stability and throughput \citep{schlag2021lineartransformerssecretlyfast}.  A parallel line scales recurrent networks themselves: RWKV introduces a receptance-weighted key–value recurrence that matches Transformer perplexity while keeping a constant-size cache \citep{peng2023rwkvreinventingrnnstransformer}; RetNet adds exponential decay for longer retention \citep{sun2023retentivenetworksuccessortransformer}; Mamba and its successor Mamba-2 treat input-conditioned state-space models as RNNs, closing most perplexity gaps on open benchmarks with linear-time generation \citep{gu2024mambalineartimesequencemodeling, dao2024mamba2}.  Gated extensions (HGRN2, Gated DeltaNet) further refine selective forgetting with per-token gates \citep{qin2024hgrn2gatedlinearrnns,yang2025gateddeltanetworksimproving}.  Despite these gains, purely linear/recurrent models still underperform Transformers on long-context retrieval and in-context learning \citep{waleffe2024empirical,parkcan,chen2024stuffed}, motivating hybrid designs. See Table \ref{tab:gen-comparison} for a more detailed overview of these models.

\paragraph{Hybrid architectures.}  
The dominant remedy is to interleave a small number of full-attention layers with many linear-time layers, yielding Transformer-like recall at a fraction of the KV-cache cost. Small-scale hybrid models include Hybrid-H3 \cite{fu2023hungry} using only two attention layers with up to 2.7B parameters, StripedHyena \cite{stripedhyena} using a 1:1 ratio with 7B parameters, RecurrentGemma \cite{recurrentgemma} using a 2:1 ratio and sliding window attention with 2B and 9B parameters, and Mamba-2-Hybrid \cite{waleffe2024empirical} using a 4:1 ratio with 8B parameters. Large-scale demonstrations include Jamba (7:1 Mamba:Transformer) \citep{lieber2024jamba,team2024jamba} and production models like MiniMax that converge on a 6–7:1 ratio \citep{minimax2025minimax01scalingfoundationmodels}. Variants trade capacity for memory: Zamba shares a single global-attention block across many recurrent blocks \citep{glorioso2024zamba,glorioso2024zamba2}; Samba keeps complexity strictly linear by pairing Mamba with sliding-window attention (SWA) \citep{ren2024sambasimplehybridstate}. Beyond block-wise mixing, \emph{prefill–decode} hybrids such as YOCO compress all keys during an $O(L)$ prefill pass and reuse them during decoding, slashing cache by $\times$layer-count \citep{sun2025you,goldstein2024goldfinch}. A finer granularity is \emph{head-wise} mixing: Hymba allocates some attention heads to softmax and the rest to state-space updates within the same layer, halving cache while preserving accuracy \citep{dong2024hymba}. Empirically, long-context quality rises steeply once a few full-attention blocks or heads are present, after which perplexity plateaus; thus the linear:full ratio primarily controls recall, whereas language modeling loss is comparatively insensitive.

\paragraph{Principled hybrid architecture design}
While these empirical studies demonstrate the viability of hybrid approaches, recent work has sought more principled design methodologies. The STAR framework \citep{thomas2025star} uses a unified mathematical foundation for linear attention mechanisms to enable automated architecture synthesis through an evolutionary optimization method. Complementarily, Poli et al. \citep{poli2024mechanistic} propose using synthetic tasks to isolate specific capabilities and understand component interactions, finding that optimal architectures leverage specialized layers through hybrid topology. These frameworks suggest that hybrid design can move beyond empirical ratio exploration toward more grounded optimization of component synergies.

\begin{figure*}[t]
    \centering
    \includegraphics[width=14cm]{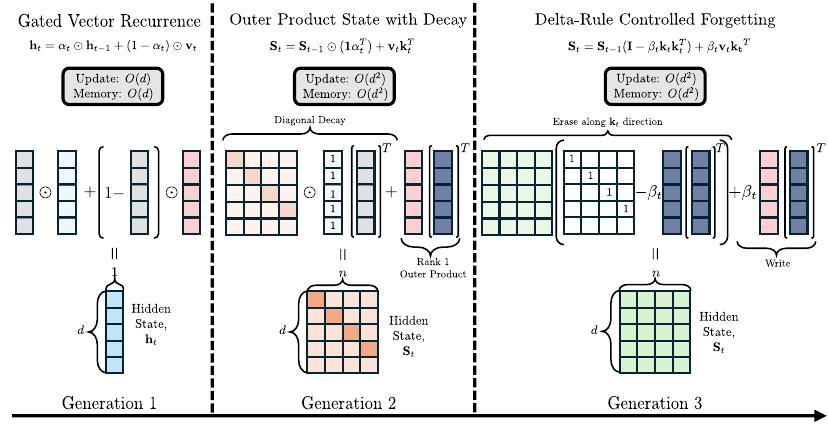}
    \caption{Three `generations' of linear-attention state updates. Generation 1 (left): Gated Vector Recurrence keeps a single vector $h_t\in\mathbb R^{d}$.  Each step first decays the previous state by an element-wise gate $\alpha_t$ (grey), then adds the gated input $v_t$ (pink).  Update and memory are both $O(d)$. Generation 2 (centre): Outer-Product State with Decay stores a full matrix $S_t\in\mathbb R^{d\times n}$.  A diagonal decay mask $(\mathbf 1\,\alpha_t^{\!\top})$ scales every column, after which a rank-1 outer product $v_t k_t^{\!\top}$ is written (pink/blue).  Cost rises to $O(d^{2})$. Generation 3 (right): $\Delta$-Rule Controlled Forgetting first erases the slice of $S_{t-1}$ aligned with key $k_t$ using the projector $I-\beta_t k_t k_t^{\!\top}$ (cells with $1$), then writes the new outer product $\beta_t v_t k_t^{\!\top}$.  This selective forget-then-write preserves capacity while matching the $O(d^{2})$ compute and memory of Generation 2. The arrow at the bottom emphasizes the historical progression from minimal to maximal recall capability.}
    \label{fig:gen123}
\end{figure*}

\section{Method}
\subsection{Linear Attention and its variants}

This section formalizes the three generations of linear–time attention mechanisms that motivate our study and explains why we benchmark the specific representatives
HGRN (first generation) and HGRN-2, GLA, RetNet (second generation).
For each generation we summarize the core update rule,
highlight the architectural property that distinguishes it from its predecessors,
and discuss its implications for hybrid design.  Table~\ref{tab:gen-comparison}
gives a unified algebraic view of the mechanisms introduced below.

\subsection{From vector recurrence to controlled forgetting}
\paragraph{Generation 1 – gated vector recurrence.}
Early linear models collapsed the per-token key–value cache into a single $d$-dimensional vector
that is updated additively and modulated by a learned gate.
Formally,
\(
\vh_t=\balpha_t\odot\vh_{t-1} + (1-\balpha_t)\odot\vv_t,\;
\vo_t=\vh_t\odot\vq_t
\)  (HGRN; \!
The element-wise gate $\balpha_t=f_\theta(x_{1:t})\in(0,1)^d$ enables
selective retention without the vanishing gradients that plagued classical RNNs,
yet the scalar state vector still struggles to store multiple competing memories.

\paragraph{Generation 2 – outer-product state with decay.}
To increase capacity, second-generation architectures promote the hidden state to a
matrix $\mathrm{S}_t\in\mathbb{R}^{d\times n}$, typically $n=d$,
accumulating rank-one outer products while applying a decay gate:
\(
\mathrm{S}_t=\mathrm{S}_{t-1}\odot(\mathbf1\balpha_t^{\!\top})+\vv_t\vk_t^{\!\top},\;
\vo_t=\mathrm{S}_t\vq_t.
\)
The form of the decay distinguishes three representative models.
\textbf{(i)~HGRN-2} ties the gate across keys and values, i.e.\ $\balpha_t=\alpha_t\mathbf1$,
yielding a hierarchical separation between slowly changing coarse memory and rapidly updated token detail \citep{qin2024hgrn2gatedlinearrnns}.  
\textbf{(ii)~GLA} lets $\balpha_t$ be fully data-dependent, learned per time-step and per channel, providing maximal flexibility but higher parameter cost \citep{yang_gated_linear_attention}.  
\textbf{(iii)~RetNet} fixes $\balpha_t=\gamma$ with $\gamma\in(0,1)$ learned once and shared across positions, giving the simplest decay and the fastest inference but no content adaptivity \citep{sun2023retentivenetworksuccessortransformer}.

\paragraph{Generation 3 – delta-rule controlled forgetting.}
The third generation abandons per-channel (diagonal) decay and instead applies a rank-1 dense transition that first erases any content aligned with the current key and then writes the new association:
\[
\mathrm{S}_t=\mathrm{S}_{t-1}\bigl(\mathbf I-\beta_t\vk_t\vk_t^{\!\top}\bigr)+\beta_t\vv_t\vk_t^{\!\top},
\qquad
\beta_t\!\in\!(0,1).
\]
Because the projector \(\mathbf I-\beta_t\vk_t\vk_t^{\!\top}\) couples all channels, information can flow across dimensions, alleviating state crowding and markedly boosting long-range recall.  The update is mathematically identical to a single stochastic-gradient step on an online least-squares objective—hence the name “delta rule’’—the hidden state behaves similar to a fast, continually trained associative memory.  This interpretation also clarifies why models such as DeltaNet and Gated DeltaNet \citep{yang2025parallelizinglineartransformersdelta,yang2025gateddeltanetworksimproving} can match Transformer perplexity and exceed it on recall-heavy tasks: each rank-1 correction injects quadratic feature interactions, giving the network a nonlinear expressiveness that standard soft-max attention (a TC\(^0\) circuit) lacks, yet without sacrificing the \(O(L)\) time and memory footprint.

\subsection{Why these representatives?}
Our main objective is to understand the roles of state size and state \mbox{management} in hybrid stacks.  
For analysis, we pick one representative for each linear attention family
and treat it as the “spokesperson" for a cluster of closely related variants:

\begin{itemize}
  \item \textbf{HGRN} (Gen-1) – the strongest vector-state model on language tasks,
        providing a clean, low-capacity baseline against which all matrix variants can be
        compared.

  \item \textbf{RetNet} (Gen-2, fixed decay) – its single scalar decay factor
        $\gamma$ is shared across positions and channels.
        Lightning uses the identical mechanism, so RetNet serves as the
        representative for all fixed-gate outer-product models.

  \item \textbf{GLA} (Gen-2, data-dependent gates) – applies a fully learned
        per-token diagonal gate.
        Mamba-2 reduces this to a data-dependent scalar gate, and
        RWKV-6 \citep{peng2023rwkvreinventingrnnstransformer} modifies only the read-out while keeping the same gated update;
        all three therefore live on the same design axis, with GLA chosen as the
        most expressive exemplar.

  \item \textbf{HGRN-2} (Gen-2, tied gate / hierarchical) – couples keys and
        values through a single gate, realizing a two-scale hierarchy at minimal cost.
        MetaLA adopts the identical tied-gate update, so HGRN-2 stands in for this class.

\end{itemize}

These four models span the spectrum from fixed through data-dependent
to hierarchically shared gating, giving us a controlled way to study how
each strategy interacts with the sparse set of full-attention layers in a hybrid model.
Generation-3 delta-rule networks (DeltaNet, Gated DeltaNet) are surveyed only briefly,
as they are currently the sole public implementations of their family; this choice
keeps our core comparison focused on the gating–versus–state-size trade-offs
within Generations 1–2.

\begin{table*}[!t]
\vspace{-2mm}
\begin{center}
\footnotesize
\begin{tabular}{l|cc|cc|cc|c}
\toprule
\textbf{Model} & \textbf{ARC-c} $\uparrow$ & \textbf{ARC-e} $\uparrow$ & \textbf{Hella} $\uparrow$ & \textbf{LMB }$\uparrow $ & \textbf{OBQA} $\uparrow$ & \textbf{PIQA} $\uparrow$ & \textbf{Avg} $\uparrow$ \\
\midrule
DeltaNet 6-1&0.299&0.594&0.410&0.334&0.334&0.675&0.441\\
GatedDeltaNet pure&0.291&0.602&0.415&0.348&0.338&0.661&0.442\\
GLA 6-1&0.287&0.588&0.421&0.343&0.334&0.671&0.441\\
HGRN 6-1&0.298&0.601&0.433&0.339&0.340&0.678&0.448\\
HGRN2 6-1&\textbf{0.300}&\textbf{0.617}&\textbf{0.436}&0.347&0.346&\textbf{0.690}&\textbf{0.456}\\
RetNet 24-1&0.287&0.574&0.411&0.309&\textbf{0.364}&0.672&0.436\\
Transformers&0.276&0.612&0.424&\textbf{0.353}&0.330&0.670&0.444\\

\bottomrule
\end{tabular}
\vspace{2mm}
\caption{340M model comparison of the best performing hybrid ratios from each linear attention variant including the transformer for zero-shot accuracy. Up arrows indicate that a higher number is better while down arrows indicate a lower number is better. 340M models show insignificant recall capability due to their relatively small parameter count. We omit recall benchmarks for this reason.}
\label{tab:340MLMComparison}
\end{center}
\end{table*}

\begin{table*}[!htbp]
\begin{center}
\footnotesize
\begin{tabular}{l|cc|cc|cc|c}
\toprule
\textbf{Model} & \textbf{ARC-c $\uparrow$} & \textbf{ARC-e $\uparrow$} & \textbf{Hella $\uparrow$} & \textbf{LMB $\uparrow$} & \textbf{OBQA $\uparrow$} & \textbf{PIQA $\uparrow$} & \textbf{Avg $\uparrow$} \\
\midrule
DeltaNet 6-1 &0.403 &0.730 &0.586 &0.475 &0.384 &\textbf{0.741} &0.553 \\
GatedDeltaNet 24-1 &\textbf{0.410} &\textbf{0.733} &0.598 &\textbf{0.502} &0.406 &\textbf{0.741} &\textbf{0.565} \\
GLA 3-1 &0.398 &0.721 &0.586 &0.488 &0.388 &0.725 &0.551 \\
HGRN 6-1 &0.392 &0.732 &0.602 &0.482 &0.412 &0.733 &0.559 \\
HGRN2 6-1 &0.404 &0.729 &\textbf{0.604} &0.500 &\textbf{0.420} &0.733 &\textbf{0.565} \\
RetNet 6-1 &0.380 &0.718 &0.584 &0.475 &0.400 &0.727 &0.547 \\
Transformers &0.395 &0.704 &0.589 &0.474 &0.400 &0.729 &0.548 \\
\bottomrule
\end{tabular}
\vspace{2mm}
\caption{1.3B parameter equivalent of Table \ref{tab:340MLMComparison}}
\label{tab:1.3BLMComparison}
\end{center}
\end{table*}

\subsection{Benchmarking framework}

For our comparative analysis, we benchmark a selection of hybridized linear attention mechanisms encompassing all three generations on language understanding and recall tasks. Specifically, we evaluate model performance on the following benchmarks: ARC-Challenge (ARC-c) \cite{clark2018thinksolvedquestionanswering}, ARC-Easy (ARC-e) \cite{clark2018thinksolvedquestionanswering}, HellaSwag \cite{zellers2019hellaswagmachinereallyfinish}, LAMBADA (LMB) \cite{paperno2016lambadadatasetwordprediction}, OpenBookQA (OBQA) \cite{mihaylov2018suitarmorconductelectricity}, and PIQA \cite{bisk2019piqareasoningphysicalcommonsense}. For recall-specific evaluation, we additionally use RULER \cite{hsieh2024rulerwhatsrealcontext}, a suite of benchmarks that assess a model's performance on retrieval, multi-hop tracing, aggregation, and question answering.

To systematically evaluate hybridization trade-offs, we test five configurations for each linear attention mechanism: linear-to-full attention ratios of 24:1, 12:1, 6:1, 3:1, and a fully linear (pure) variant. As a baseline, we also include a standard full-transformer model.

All models are pretrained on the fineweb-edu~\cite{penedo2024fineweb} dataset and using flash-linear-attention library~\cite{yang2024fla}. The 340M-parameter models are trained on 20 billion tokens, and the 1.3B-parameter models are trained on 100 billion tokens. All models are optimized using the AdamW optimizer with a cosine learning rate schedule. For the 340M-parameter models, we use a batch size of 50k tokens, accumulating a total of 20 billion training tokens. For the 1.3B-parameter models, the batch size is set to 1 million tokens, reaching a total of 100 billion training tokens. All evaluation benchmarks are conducted in a zero-shot manner without any task-specific fine-tuning or prompt engineering.

\section{Empirical Study}
\label{sec:emprical}
The empirical analysis then unfolds in three steps: 
\textbf{First}, we compare each linear-attention backbone in isolation and in a hybridized form that interleaves full-attention layers, revealing whether strong standalone results translate into strong hybrids.  
\textbf{Next}, keeping all other factors fixed, we vary the linear-to-full ratio (24:1, 12:1, 6:1, 3:1, and the pure linear case) and observe its effect on language-model accuracy and long-range recall as measured by RULER.  
\textbf{Finally}, by contrasting model families that differ in gating, recurrence hierarchy, and forgetting mechanisms, we identify the architectural ingredients that let hybrids match Transformer-level recall with a much smaller KV cache.

\subsection{Standalone Performance versus Hybrid Performance}

We investigate whether linear-attention architectures that excel in standalone evaluation maintain their relative performance when combined with full attention, in hybrid configurations. Tables \ref{tab:340MLMComparison} and \ref{tab:1.3BLMComparison} show that this is not the case.

At the 340M parameter scale, GatedDeltaNet achieves the highest standalone accuracy. However, when hybridized, HGRN2 at a 6:1 linear-to-full attention ratio outperforms it, exceeding both the Transformer baseline and the standalone leader by 1.2 percentage points. 

The same pattern emerges at 1.3B parameters: GatedDeltaNet leads in standalone evaluation but becomes comparable with HGRN2 after hybridization. GatedDeltaNet at 24:1 and HGRN2 at 6:1 achieve equivalent performance, with several other hybrid configurations within one percentage point.

The relative performance of different architectures changes substantially when linear and full attention mechanisms are combined, making standalone scores unreliable predictors of hybrid performance.

\subsection{Impact of the Linear-to-Full Attention Ratio}
\begin{figure*}
    \centering
    \includegraphics[width=0.8\linewidth]{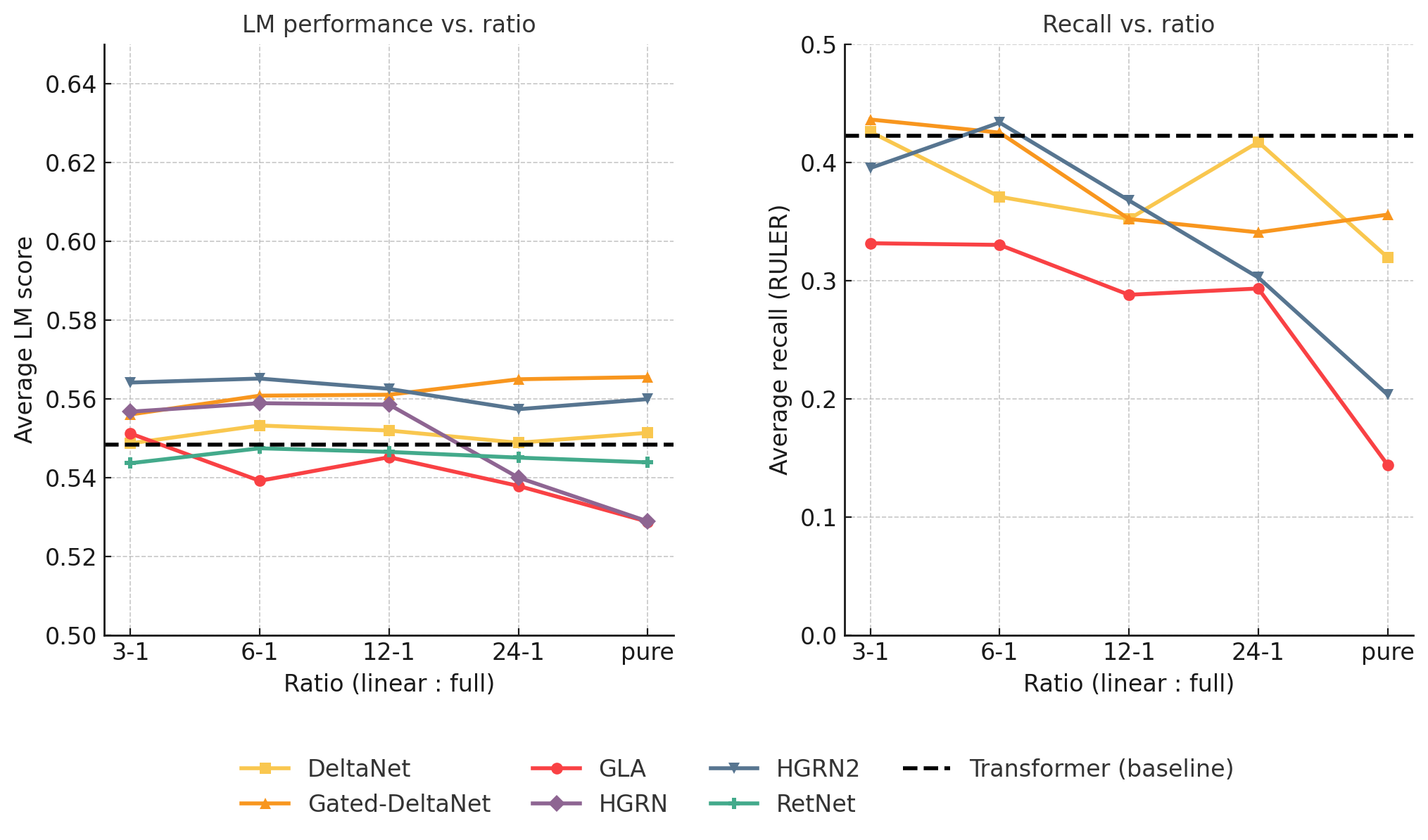}
    \caption{Language performance and recall performance tasks are averaged and compared over varying ratios}
    \label{fig:trend-graph}
\end{figure*}

We examine how the linear-to-full attention ratio affects language modeling and recall performance. Figure \ref{fig:trend-graph} shows the behavior of these two capabilities across different architectures and ratios. On the left panel, language modeling performance remains largely flat across all ratio configurations. Most architectures cluster around 0.55-0.57 average LM score, with minimal variation. A tabular version of both graphs can be found in Appendix \ref{sec:TableRatioResults}

The right panel reveals a markedly different pattern for recall performance. All architectures show an upward trend as the proportion of full-attention layers increases. Performance rises from pure linear configurations (around 0.1-0.35 RULER score) toward the full-attention baseline (dashed line at approximately 0.42). Notably, most architectures approach or exceed this baseline at the 3:1 ratio, with some like DeltaNet and Gated-DeltaNet achieving their peak recall performance at this configuration. This trend demonstrates that recall capability benefits substantially from increased full-attention allocation, unlike language modeling performance. This trend is likely driven by the effectively unbounded hidden state provided by the KV cache. As the number of full attention layers increases, hybrid models will tend to exhibit stronger performance on recall tasks.

The behavior of these two capabilities has practical implications. Systems prioritizing language modeling can operate efficiently with high linear-to-full ratios, while applications requiring long-range recall need more balanced attention distributions.

\subsection{Architectural Determinants of Hybrid Effectiveness}
\label{sec:hybrid-arch}

Our experiments indicate that hybrid performance hinges on three complementary architectural properties rather than any single mechanism. Figure~\ref{fig:trend-graph} provides a visual backdrop; the analysis below is necessarily observational. Ablations are left to future work.

\paragraph{Selective gating prevents catastrophic overwriting.}
Architectures that expose their hidden state to a learned, token-wise gate—GatedDeltaNet and HGRN-2—consistently perform best in recall once hybridised, surpassing the Transformer baseline by 2–5 percentage points at the optimal ratio.  By contrast, RetNet’s fixed exponential decay fails to protect long-range cues, yielding near-zero recall even when full-attention layers are added.

\paragraph{Hierarchical recurrence supplies multi-timescale context.}
The two-level pathway in HGRN-2 stores coarse summaries at a slower update rate while the fast path handles token-level details.  Relative to its single-path ablation (HGRN), this hierarchy doubles recall and improves the LM–recall trade-off in Figure~\ref{fig:trend-graph}, suggesting that widely spaced full-attention layers benefit from a recurrent hierarchy that can “latch’’ information between them.

\paragraph{Controlled forgetting curbs state crowding.}
GatedDeltaNet realises controlled forgetting with an outer-product delta rule,
whereas HGRN-2 achieves the same goal through gated diagonal decay. Although only the former subtracts stale content explicitly, both mechanisms prevent the unbounded accumulation that plagues purely additive updates (e.g.\ GLA) and therefore attain strong recall scores.

\paragraph{Interaction with the linear to full ratio.}
Language-model accuracy varies by less than one percent across ratios, but recall rises steadily as full-attention layers are added and saturates around 3:1.  Architectures lacking either gated or delta-style forgetting never reach Transformer-level recall, regardless of ratio, implying that a suitable model architecture is necessary to attain results on par with the Transformer.

\paragraph{Practical guideline.}
A memory-constrained practitioner can select a gated, hierarchically recurrent model with controlled forgetting—concretely, HGRN-2 or GatedDeltaNet—and devote roughly one quarter of the layers to full attention.  In our 1.3B-parameter setting this configuration achieved near-Transformer recall while cutting KV-cache memory by a factor of four.  Verifying the trend at larger scales remains future work.

\subsection{Summary of Findings and Practical Guidelines}\label{sec:summary}
\textit{This section condenses our empirical study into clear, actionable insights for building memory‑efficient long‑context language models.}

\begin{itemize}
\item \textbf{Hybrid quality cannot be inferred from standalone performance.} GatedDeltaNet is strongest in purely linear form, yet HGRN‑2 delivers the best results once full‑attention layers are added, showing that standalone benchmarks are insufficient when selecting a hybrid backbone.
\item \textbf{Recall—not perplexity—determines the optimal linear:full mix.} Moving from a 24:1 to a 3:1 ratio nearly doubles RULER recall while shifting language‑model loss by less than 1, indicating that memory‑centric metrics should drive architectural choices.
\item \textbf{Three architectural ingredients underpin effective hybrids.} Selective gating, hierarchical recurrence, and controlled forgetting jointly enable Transformer‑level recall with a small KV cache; omitting any one of these components degrades retrieval markedly.
\item \textbf{Recommended deployment recipe.} Employ a gated, hierarchical backbone (e.g., HGRN‑2 or GatedDeltaNet) with one soft‑max attention layer for every 3–6 linear layers. In our 1.3B‑parameter setting this achieves near‑Transformer recall while shrinking the KV cache by a factor of 4–7.
\end{itemize}

\section{Conclusion}
\label{sec:conclusion}
We delivered the first systematic comparison of three generations of linear‑time attention mechanisms, evaluated both in isolation and within hybrid architectures that interleave a minority of full‑attention layers. The study shows that, when equipped with gating, hierarchical recurrence, and controlled forgetting, linear backbones can match Transformer‑level recall at a fraction of the KV‑cache cost; a 3:1–6:1 (linear:full) ratio provides a practical balance between memory and quality.

\section*{Limitations}

Our analysis is confined to models up to 1.3B parameters, a 4,096‑token context window, and block‑wise mixing ratios. The question of whether or not the observed trade‑offs persist at the 10B+ scale, with 128k token contexts, or under instruction‑tuning and multilingual data remains open. We will, however, point out that models like MiniMax-01 \cite{minimax2025minimax01scalingfoundationmodels} have proven that hybrid models can properly scale up to the modern language model size. Exploring finer‑grained hybrids (e.g., head‑wise or dynamic routing) and automated architecture search (e.g., the STAR framework) constitute promising directions for extending this work.

\bibliography{custom}

\begin{thebibliography}{46}
\providecommand{\natexlab}[1]{#1}

\bibitem[{Abreu et~al.(2025)Abreu, Shrestha, Zhu, and Eshraghian}]{abreu2025neuromorphic}
Steven Abreu, Sumit~Bam Shrestha, Rui-Jie Zhu, and Jason Eshraghian. 2025.
\newblock \href {https://openreview.net/forum?id=qaDM1R2nlm} {Neuromorphic principles for efficient large language models on intel loihi 2}.
\newblock In \emph{First Workshop on Scalable Optimization for Efficient and Adaptive Foundation Models}.

\bibitem[{Bisk et~al.(2019)Bisk, Zellers, Bras, Gao, and Choi}]{bisk2019piqareasoningphysicalcommonsense}
Yonatan Bisk, Rowan Zellers, Ronan~Le Bras, Jianfeng Gao, and Yejin Choi. 2019.
\newblock \href {https://arxiv.org/abs/1911.11641} {Piqa: Reasoning about physical commonsense in natural language}.
\newblock \emph{Preprint}, arXiv:1911.11641.

\bibitem[{Botev et~al.(2024)Botev, De, Smith, Fernando, Muraru, Haroun, Berrada, Pascanu, Sessa, Dadashi, Hussenot, Ferret, Girgin, Bachem, Andreev, Kenealy, Mesnard, Hardin, Bhupatiraju, Pathak, Sifre, Rivi{\`e}re, Kale, Love, Tafti, Joulin, Fiedel, Senter, Chen, Srinivasan, Desjardins, Budden, Doucet, Vikram, Paszke, Gale, Borgeaud, Chen, Brock, Paterson, Brennan, Risdal, Gundluru, Devanathan, Mooney, Chauhan, Culliton, Martins, Bandy, Huntsperger, Cameron, Zucker, Warkentin, Peran, Giang, Ghahramani, Farabet, Kavukcuoglu, Hassabis, Hadsell, Teh, and de~Frietas}]{recurrentgemma}
Aleksandar Botev, Soham De, Samuel~L. Smith, Anushan Fernando, George-Cristian Muraru, Ruba Haroun, Leonard Berrada, Razvan Pascanu, Pier~Giuseppe Sessa, Robert Dadashi, L{\'e}onard Hussenot, Johan Ferret, Sertan Girgin, Olivier Bachem, Alek Andreev, Kathleen Kenealy, Thomas Mesnard, Cassidy Hardin, Surya Bhupatiraju, and 43 others. 2024.
\newblock \href {https://arxiv.org/abs/2404.07839} {{RecurrentGemma}: Moving past transformers for efficient open language models}.
\newblock \emph{Preprint}, arXiv:2404.07839.

\bibitem[{Chen et~al.(2024)Chen, Zhang, Hu, Han, Liu, and Sun}]{chen2024stuffed}
Yingfa Chen, Xinrong Zhang, Shengding Hu, Xu~Han, Zhiyuan Liu, and Maosong Sun. 2024.
\newblock \href {https://arxiv.org/abs/2410.07145} {Stuffed mamba: State collapse and state capacity of {RNN}-based long-context modeling}.
\newblock \emph{Preprint}, arXiv:2410.07145.

\bibitem[{Chou et~al.(2024)Chou, Yao, Wang, Pan, Zhu, Wu, Zhong, Qiao, Xu, and Li}]{chou2024metala}
Yuhong Chou, Man Yao, Kexin Wang, Yuqi Pan, Rui-Jie Zhu, Jibin Wu, Yiran Zhong, Yu~Qiao, Bo~Xu, and Guoqi Li. 2024.
\newblock Metala: Unified optimal linear approximation to softmax attention map.
\newblock \emph{Advances in Neural Information Processing Systems}, 37:71034--71067.

\bibitem[{Clark et~al.(2018)Clark, Cowhey, Etzioni, Khot, Sabharwal, Schoenick, and Tafjord}]{clark2018thinksolvedquestionanswering}
Peter Clark, Isaac Cowhey, Oren Etzioni, Tushar Khot, Ashish Sabharwal, Carissa Schoenick, and Oyvind Tafjord. 2018.
\newblock \href {https://arxiv.org/abs/1803.05457} {Think you have solved question answering? try arc, the ai2 reasoning challenge}.
\newblock \emph{Preprint}, arXiv:1803.05457.

\bibitem[{Dao and Gu(2024)}]{dao2024mamba2}
Tri Dao and Albert Gu. 2024.
\newblock Transformers are ssms: Generalized models and efficient algorithms through structured state space duality.
\newblock \emph{arXiv preprint arXiv:2405.21060}.

\bibitem[{De et~al.(2024)De, Smith, Fernando, Botev, Cristian-Muraru, Gu, Haroun, Berrada, Chen, Srinivasan et~al.}]{de2024griffin}
Soham De, Samuel~L Smith, Anushan Fernando, Aleksandar Botev, George Cristian-Muraru, Albert Gu, Ruba Haroun, Leonard Berrada, Yutian Chen, Srivatsan Srinivasan, and 1 others. 2024.
\newblock Griffin: Mixing gated linear recurrences with local attention for efficient language models.
\newblock \emph{arXiv preprint arXiv:2402.19427}.

\bibitem[{Dong et~al.(2024)Dong, Fu, Diao, Byeon, Chen et~al.}]{dong2024hymba}
Xin Dong, Yonggan Fu, Shizhe Diao, Wonmin Byeon, Zijia Chen, and 1 others. 2024.
\newblock \href {https://arxiv.org/abs/2411.13676} {Hymba: A hybrid-head architecture for small language models}.
\newblock \emph{Preprint}, arXiv:2411.13676.

\bibitem[{Fu et~al.(2023)Fu, Dao, Saab, Thomas, Rudra, and Re}]{fu2023hungry}
Daniel~Y Fu, Tri Dao, Khaled~Kamal Saab, Armin~W Thomas, Atri Rudra, and Christopher Re. 2023.
\newblock \href {https://openreview.net/forum?id=COZDy0WYGg} {Hungry hungry hippos: Towards language modeling with state space models}.
\newblock In \emph{The Eleventh International Conference on Learning Representations}.

\bibitem[{Glorioso et~al.(2024{\natexlab{a}})Glorioso, Anthony, Tokpanov, Golubeva, Shyam et~al.}]{glorioso2024zamba2}
Paolo Glorioso, Quentin Anthony, Yury Tokpanov, Anna Golubeva, Vasudev Shyam, and 1 others. 2024{\natexlab{a}}.
\newblock \href {https://arxiv.org/abs/2411.15242} {The zamba2 suite: Technical report}.
\newblock \emph{Preprint}, arXiv:2411.15242.

\bibitem[{Glorioso et~al.(2024{\natexlab{b}})Glorioso, Anthony, Tokpanov, Whittington, Pilault, and Millidge}]{glorioso2024zamba}
Paolo Glorioso, Quentin Anthony, Yury Tokpanov, James Whittington, Jonathan Pilault, and Beren Millidge. 2024{\natexlab{b}}.
\newblock \href {https://arxiv.org/abs/2405.16712} {Zamba: A compact 7b ssm hybrid model}.
\newblock \emph{Preprint}, arXiv:2405.16712.

\bibitem[{Goldstein et~al.(2024)Goldstein, Goldblum, Chen, and Goldstein}]{goldstein2024goldfinch}
Elliott Goldstein, Michael Goldblum, Minshuo Chen, and Tom Goldstein. 2024.
\newblock \href {https://arxiv.org/abs/2407.12077} {Goldfinch: High performance rwkv/transformer hybrid with linear pre-fill and extreme kv-cache compression}.
\newblock \emph{Preprint}, arXiv:2407.12077.

\bibitem[{Gu and Dao(2024)}]{gu2024mambalineartimesequencemodeling}
Albert Gu and Tri Dao. 2024.
\newblock \href {https://arxiv.org/abs/2312.00752} {Mamba: Linear-time sequence modeling with selective state spaces}.
\newblock \emph{Preprint}, arXiv:2312.00752.

\bibitem[{Hsieh et~al.(2024)Hsieh, Sun, Kriman, Acharya, Rekesh, Jia, Zhang, and Ginsburg}]{hsieh2024rulerwhatsrealcontext}
Cheng-Ping Hsieh, Simeng Sun, Samuel Kriman, Shantanu Acharya, Dima Rekesh, Fei Jia, Yang Zhang, and Boris Ginsburg. 2024.
\newblock \href {https://arxiv.org/abs/2404.06654} {Ruler: What's the real context size of your long-context language models?}
\newblock \emph{Preprint}, arXiv:2404.06654.

\bibitem[{{Jamba Team}(2024)}]{team2024jamba}
{Jamba Team}. 2024.
\newblock \href {https://arxiv.org/abs/2408.12570} {Jamba-1.5: Hybrid transformer--mamba models at scale}.
\newblock \emph{Preprint}, arXiv:2408.12570.

\bibitem[{Jelassi et~al.(2024)Jelassi, Brandfonbrener, Kakade, and Malach}]{jelassi2024repeat}
Samy Jelassi, David Brandfonbrener, Sham~M. Kakade, and Eran Malach. 2024.
\newblock Repeat after me: transformers are better than state space models at copying.
\newblock In \emph{Proceedings of the 41st International Conference on Machine Learning}, ICML'24. JMLR.org.

\bibitem[{Katharopoulos et~al.(2020)Katharopoulos, Vyas, Pappas, and Fleuret}]{katharopoulos2020transformersrnnsfastautoregressive}
Angelos Katharopoulos, Apoorv Vyas, Nikolaos Pappas, and François Fleuret. 2020.
\newblock \href {https://arxiv.org/abs/2006.16236} {Transformers are rnns: Fast autoregressive transformers with linear attention}.
\newblock \emph{Preprint}, arXiv:2006.16236.

\bibitem[{Lieber et~al.(2024)Lieber, Lenz, Bata, Cohen, Osin, Dalmedigos, Safahi, Meirom, Belinkov, {Shalev-Shwartz}, Abend, Alon, Asida, Bergman, Glozman, Gokhman, Manevich, Ratner, Rozen, Shwartz, Zusman, and Shoham}]{lieber2024jamba}
Opher Lieber, Barak Lenz, Hofit Bata, Gal Cohen, Jhonathan Osin, Itay Dalmedigos, Erez Safahi, Shaked Meirom, Yonatan Belinkov, Shai {Shalev-Shwartz}, Omri Abend, Raz Alon, Tomer Asida, Amir Bergman, Roman Glozman, Michael Gokhman, Avashalom Manevich, Nir Ratner, Noam Rozen, and 3 others. 2024.
\newblock \href {https://arxiv.org/abs/2403.19887} {Jamba: A hybrid {Transformer}--{Mamba} language model}.
\newblock \emph{Preprint}, arXiv:2403.19887.

\bibitem[{Liu et~al.(2025)Liu, Gao, and Chen}]{liu2025scaling}
Kai Liu, Jianfei Gao, and Kai Chen. 2025.
\newblock Scaling up the state size of rnn llms for long-context scenarios.
\newblock In \emph{Proceedings of the 63rd Annual Meeting of the Association for Computational Linguistics}. Association for Computational Linguistics.

\bibitem[{Mihaylov et~al.(2018)Mihaylov, Clark, Khot, and Sabharwal}]{mihaylov2018suitarmorconductelectricity}
Todor Mihaylov, Peter Clark, Tushar Khot, and Ashish Sabharwal. 2018.
\newblock \href {https://arxiv.org/abs/1809.02789} {Can a suit of armor conduct electricity? a new dataset for open book question answering}.
\newblock \emph{Preprint}, arXiv:1809.02789.

\bibitem[{{MiniMax AI Team}(2025)}]{minimax2025minimax01scalingfoundationmodels}
{MiniMax AI Team}. 2025.
\newblock \href {https://arxiv.org/abs/2501.08313} {Minimax-01: Scaling foundation models with lightning attention}.
\newblock \emph{Preprint}, arXiv:2501.08313.

\bibitem[{Paperno et~al.(2016)Paperno, Kruszewski, Lazaridou, Pham, Bernardi, Pezzelle, Baroni, Boleda, and Fernández}]{paperno2016lambadadatasetwordprediction}
Denis Paperno, Germán Kruszewski, Angeliki Lazaridou, Quan~Ngoc Pham, Raffaella Bernardi, Sandro Pezzelle, Marco Baroni, Gemma Boleda, and Raquel Fernández. 2016.
\newblock \href {https://arxiv.org/abs/1606.06031} {The lambada dataset: Word prediction requiring a broad discourse context}.
\newblock \emph{Preprint}, arXiv:1606.06031.

\bibitem[{Park et~al.(2024)Park, Park, Xiong, Lee, Cho, Oymak, Lee, and Papailiopoulos}]{parkcan}
Jongho Park, Jaeseung Park, Zheyang Xiong, Nayoung Lee, Jaewoong Cho, Samet Oymak, Kangwook Lee, and Dimitris Papailiopoulos. 2024.
\newblock Can mamba learn how to learn? a comparative study on in-context learning tasks.
\newblock \emph{arXiv preprint arXiv:2402.04248}.

\bibitem[{Penedo et~al.(2024)Penedo, Kydl{\'\i}{\v{c}}ek, Lozhkov, Mitchell, Raffel, Von~Werra, Wolf et~al.}]{penedo2024fineweb}
Guilherme Penedo, Hynek Kydl{\'\i}{\v{c}}ek, Anton Lozhkov, Margaret Mitchell, Colin~A Raffel, Leandro Von~Werra, Thomas Wolf, and 1 others. 2024.
\newblock The fineweb datasets: Decanting the web for the finest text data at scale.
\newblock \emph{Advances in Neural Information Processing Systems}, 37:30811--30849.

\bibitem[{Peng et~al.(2023)Peng, Alcaide, Anthony, Albalak, Arcadinho, Biderman, Cao, Cheng, Chung, Grella, GV, He, Hou, Lin, Kazienko, Kocon, Kong, Koptyra, Lau, Mantri, Mom, Saito, Song, Tang, Wang, Wind, Wozniak, Zhang, Zhang, Zhao, Zhou, Zhou, Zhu, and Zhu}]{peng2023rwkvreinventingrnnstransformer}
Bo~Peng, Eric Alcaide, Quentin Anthony, Alon Albalak, Samuel Arcadinho, Stella Biderman, Huanqi Cao, Xin Cheng, Michael Chung, Matteo Grella, Kranthi~Kiran GV, Xuzheng He, Haowen Hou, Jiaju Lin, Przemyslaw Kazienko, Jan Kocon, Jiaming Kong, Bartlomiej Koptyra, Hayden Lau, and 15 others. 2023.
\newblock \href {https://arxiv.org/abs/2305.13048} {Rwkv: Reinventing rnns for the transformer era}.
\newblock \emph{Preprint}, arXiv:2305.13048.

\bibitem[{Peng et~al.(2024)Peng, Goldstein, Anthony, Albalak, Alcaide, Biderman, Cheah, Du, Ferdinan, Hou, Kazienko, GV, Kocoń, Koptyra, Krishna, Jr., Lin, Muennighoff, Obeid, Saito, Song, Tu, Wirawan, Woźniak, Zhang, Zhao, Zhao, Zhou, Zhu, and Zhu}]{peng2024eaglefinchrwkvmatrixvalued}
Bo~Peng, Daniel Goldstein, Quentin Anthony, Alon Albalak, Eric Alcaide, Stella Biderman, Eugene Cheah, Xingjian Du, Teddy Ferdinan, Haowen Hou, Przemysław Kazienko, Kranthi~Kiran GV, Jan Kocoń, Bartłomiej Koptyra, Satyapriya Krishna, Ronald~McClelland Jr., Jiaju Lin, Niklas Muennighoff, Fares Obeid, and 11 others. 2024.
\newblock \href {https://arxiv.org/abs/2404.05892} {Eagle and finch: Rwkv with matrix-valued states and dynamic recurrence}.
\newblock \emph{Preprint}, arXiv:2404.05892.

\bibitem[{Poli et~al.(2024)Poli, Thomas, Nguyen, Ponnusamy, Deiseroth, Kersting, Suzuki, Hie, Ermon, R\'{e}, Zhang, and Massaroli}]{poli2024mechanistic}
Michael Poli, Armin~W Thomas, Eric Nguyen, Pragaash Ponnusamy, Bj\"{o}rn Deiseroth, Kristian Kersting, Taiji Suzuki, Brian Hie, Stefano Ermon, Christopher R\'{e}, Ce~Zhang, and Stefano Massaroli. 2024.
\newblock Mechanistic design and scaling of hybrid architectures.
\newblock In \emph{Proceedings of the 41st International Conference on Machine Learning}, ICML'24.

\bibitem[{Poli et~al.(2023)Poli, Wang, Massaroli, Quesnelle, Carlow, Nguyen, and Thomas}]{stripedhyena}
Michael Poli, Jue Wang, Stefano Massaroli, Jeffrey Quesnelle, Ryan Carlow, Eric Nguyen, and Armin Thomas. 2023.
\newblock \href {https://doi.org/10.57967/hf/1595} {Stripedhyena: Moving beyond transformers with hybrid signal processing models}.

\bibitem[{Qin et~al.(2024)Qin, Yang, Sun, Shen, Li, Sun, and Zhong}]{qin2024hgrn2gatedlinearrnns}
Zhen Qin, Songlin Yang, Weixuan Sun, Xuyang Shen, Dong Li, Weigao Sun, and Yiran Zhong. 2024.
\newblock \href {https://openreview.net/forum?id=y6SqbJfCSk} {{HGRN2}: Gated linear {RNN}s with state expansion}.
\newblock In \emph{First Conference on Language Modeling}.

\bibitem[{Ren et~al.(2024)Ren, Liu, Lu, Shen, Liang, and Chen}]{ren2024sambasimplehybridstate}
Liliang Ren, Yang Liu, Yadong Lu, Yelong Shen, Chen Liang, and Weizhu Chen. 2024.
\newblock \href {https://arxiv.org/abs/2406.07522} {Samba: Simple hybrid state space models for efficient unlimited context language modeling}.
\newblock \emph{Preprint}, arXiv:2406.07522.

\bibitem[{Schlag et~al.(2021)Schlag, Irie, and Schmidhuber}]{schlag2021lineartransformerssecretlyfast}
Imanol Schlag, Kazuki Irie, and J{\"u}rgen Schmidhuber. 2021.
\newblock Linear transformers are secretly fast weight programmers.
\newblock In \emph{Proceedings of the 38th International Conference on Machine Learning}, volume 139. PMLR.

\bibitem[{Shen et~al.(2024)Shen, Li, Leng, Qin, Sun, and Zhong}]{shen2024scaling}
Xuyang Shen, Dong Li, Ruitao Leng, Zhen Qin, Weigao Sun, and Yiran Zhong. 2024.
\newblock Scaling laws for linear complexity language models.
\newblock \emph{arXiv preprint arXiv:2406.16690}.

\bibitem[{Sieber et~al.(2024)Sieber, Alonso, Didier, Zeilinger, and Orvieto}]{sieber2024understanding}
Jerome Sieber, Carmen~Amo Alonso, Alexandre Didier, Melanie Zeilinger, and Antonio Orvieto. 2024.
\newblock \href {https://openreview.net/forum?id=iF7MnXnxRw} {Understanding the differences in foundation models: Attention, state space models, and recurrent neural networks}.
\newblock In \emph{The Thirty-eighth Annual Conference on Neural Information Processing Systems}.

\bibitem[{Sun et~al.(2023)Sun, Dong, Huang, Ma, Xia, Xue, Wang, and Wei}]{sun2023retentivenetworksuccessortransformer}
Yutao Sun, Li~Dong, Shaohan Huang, Shuming Ma, Yuqing Xia, Jilong Xue, Jianyong Wang, and Furu Wei. 2023.
\newblock \href {https://arxiv.org/abs/2307.08621} {Retentive network: A successor to transformer for large language models}.
\newblock \emph{Preprint}, arXiv:2307.08621.

\bibitem[{Sun et~al.(2024)Sun, Dong, Zhu, Huang, Wang, Ma, Zhang, Wang, and Wei}]{sun2025you}
Yutao Sun, Li~Dong, Yi~Zhu, Shaohan Huang, Wenhui Wang, Shuming Ma, Quanlu Zhang, Jianyong Wang, and Furu Wei. 2024.
\newblock \href {https://arxiv.org/abs/2405.05254} {You only cache once: Decoder-decoder architectures for language models}.
\newblock \emph{Preprint}, arXiv:2405.05254.

\bibitem[{Thomas et~al.(2025)Thomas, Parnichkun, Amini, Massaroli, and Poli}]{thomas2025star}
Armin~W Thomas, Rom Parnichkun, Alexander Amini, Stefano Massaroli, and Michael Poli. 2025.
\newblock \href {https://openreview.net/forum?id=HsHxSN23rM} {{STAR}: Synthesis of tailored architectures}.
\newblock In \emph{The Thirteenth International Conference on Learning Representations}.

\bibitem[{Vaswani et~al.(2023)Vaswani, Shazeer, Parmar, Uszkoreit, Jones, Gomez, Kaiser, and Polosukhin}]{vaswani2023attentionneed}
Ashish Vaswani, Noam Shazeer, Niki Parmar, Jakob Uszkoreit, Llion Jones, Aidan~N. Gomez, Lukasz Kaiser, and Illia Polosukhin. 2023.
\newblock \href {https://arxiv.org/abs/1706.03762} {Attention is all you need}.
\newblock \emph{Preprint}, arXiv:1706.03762.

\bibitem[{Voelker et~al.(2019)Voelker, Kaji{\'c}, and Eliasmith}]{voelker2019legendre}
Aaron Voelker, Ivana Kaji{\'c}, and Chris Eliasmith. 2019.
\newblock Legendre memory units: Continuous-time representation in recurrent neural networks.
\newblock \emph{Advances in neural information processing systems}, 32.

\bibitem[{Waleffe et~al.(2024)Waleffe, Byeon, Riach, Norick, Korthikanti, Dao, Gu, Hatamizadeh, Singh, Narayanan et~al.}]{waleffe2024empirical}
Roger Waleffe, Wonmin Byeon, Duncan Riach, Brandon Norick, Vijay Korthikanti, Tri Dao, Albert Gu, Ali Hatamizadeh, Sudhakar Singh, Deepak Narayanan, and 1 others. 2024.
\newblock An empirical study of mamba-based language models.
\newblock \emph{arXiv preprint arXiv:2406.07887}.

\bibitem[{Yang et~al.(2025{\natexlab{a}})Yang, Kautz, and Hatamizadeh}]{yang2025gateddeltanetworksimproving}
Songlin Yang, Jan Kautz, and Ali Hatamizadeh. 2025{\natexlab{a}}.
\newblock \href {https://openreview.net/forum?id=r8H7xhYPwz} {Gated delta networks: Improving {Mamba2} with delta rule}.
\newblock In \emph{The Thirteenth International Conference on Learning Representations}.

\bibitem[{Yang et~al.(2024)Yang, Wang, Shen, Panda, and Kim}]{yang_gated_linear_attention}
Songlin Yang, Bailin Wang, Yikang Shen, Rameswar Panda, and Yoon Kim. 2024.
\newblock Gated linear attention transformers with hardware-efficient training.
\newblock In \emph{Proceedings of the 41st International Conference on Machine Learning (ICML)}.

\bibitem[{Yang et~al.(2025{\natexlab{b}})Yang, Wang, Zhang, Shen, and Kim}]{yang2025parallelizinglineartransformersdelta}
Songlin Yang, Bailin Wang, Yu~Zhang, Yikang Shen, and Yoon Kim. 2025{\natexlab{b}}.
\newblock \href {https://arxiv.org/abs/2406.06484} {Parallelizing linear transformers with the delta rule over sequence length}.
\newblock \emph{Preprint}, arXiv:2406.06484.

\bibitem[{Yang and Zhang(2024)}]{yang2024fla}
Songlin Yang and Yu~Zhang. 2024.
\newblock Fla: A triton-based library for hardware-efficient implementations of linear attention mechanism.

\bibitem[{Zellers et~al.(2019)Zellers, Holtzman, Bisk, Farhadi, and Choi}]{zellers2019hellaswagmachinereallyfinish}
Rowan Zellers, Ari Holtzman, Yonatan Bisk, Ali Farhadi, and Yejin Choi. 2019.
\newblock \href {https://arxiv.org/abs/1905.07830} {Hellaswag: Can a machine really finish your sentence?}
\newblock \emph{Preprint}, arXiv:1905.07830.

\bibitem[{Zhu et~al.(2024)Zhu, Zhang, Sifferman, Sheaves, Wang, Richmond, Zhou, and Eshraghian}]{zhu2024scalablematmulfreelanguagemodeling}
Rui-Jie Zhu, Yu~Zhang, Ethan Sifferman, Tyler Sheaves, Yiqiao Wang, Dustin Richmond, Peng Zhou, and Jason~K. Eshraghian. 2024.
\newblock \href {https://arxiv.org/abs/2406.02528} {Scalable matmul-free language modeling}.
\newblock \emph{Preprint}, arXiv:2406.02528.

\end{thebibliography}

\appendix
\onecolumn

\section{Performance-Efficiency Pareto Front}
\label{sec:performance-efficiency}

\begin{figure*}[t] 
    \centering
    \begin{subfigure}[b]{0.48\linewidth}
        \centering
        \includegraphics[width=\linewidth]{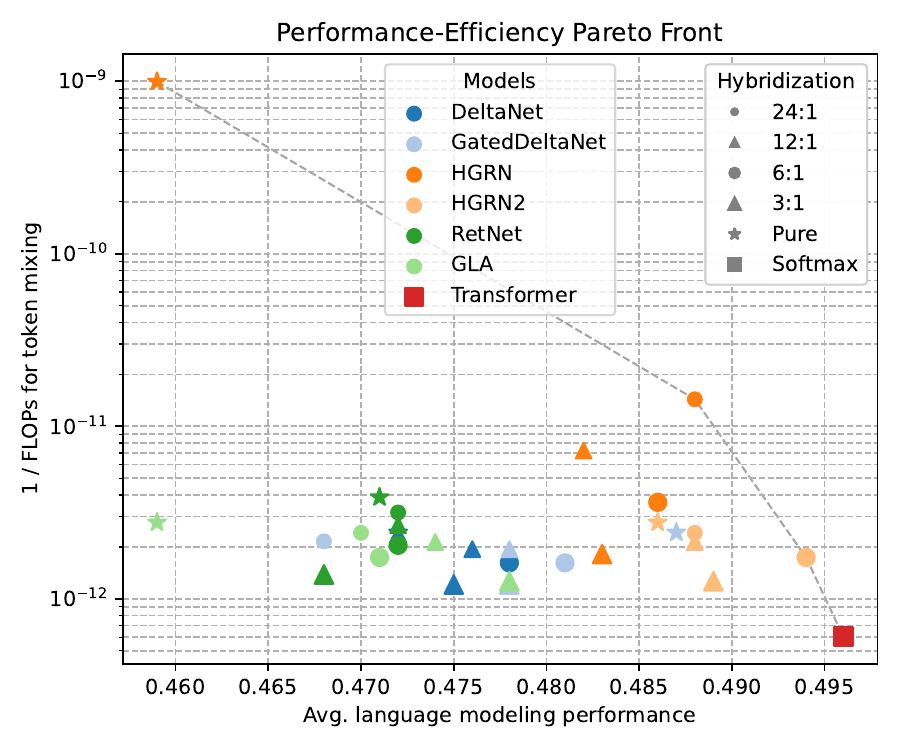}
        \caption{Sequence length $L=4096$.}
        \label{fig:pareto-front-4k}
    \end{subfigure}
    \hfill 
    \begin{subfigure}[b]{0.48\linewidth}
        \centering
        \includegraphics[width=\linewidth]{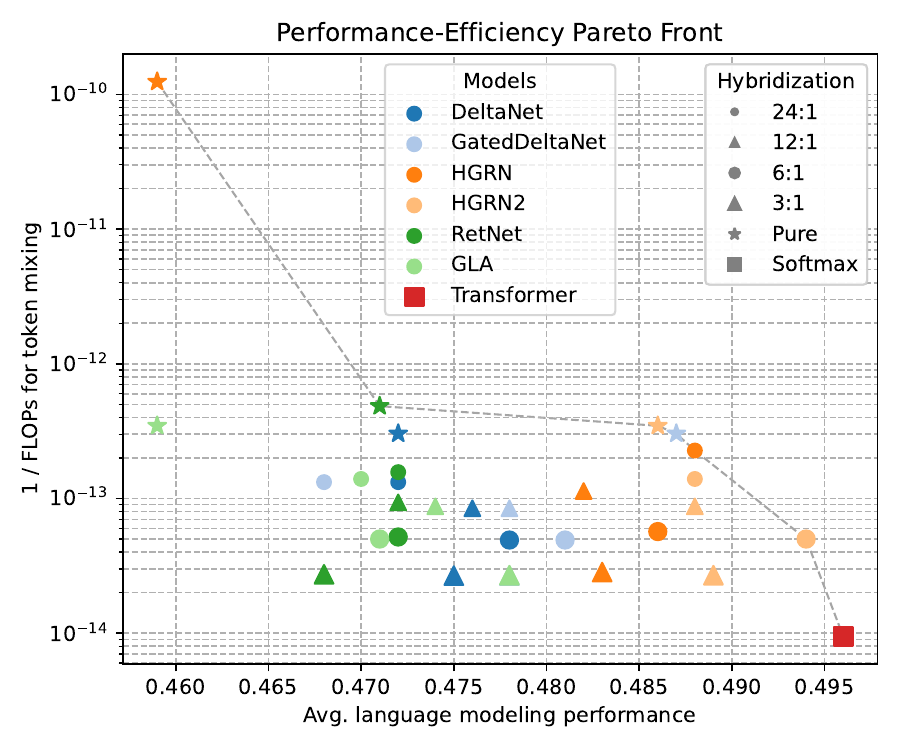}
        \caption{Sequence length $L=32768$.}
        \label{fig:pareto-front-32k}
    \end{subfigure}
    \caption{Performance-Efficiency Pareto Front where FLOPs are calculated for the 1.3B parameter models with two different sequence lengths. Note that the y-axis is inverted so that the Pareto Front can be clearly seen towards the top-right.}
    \label{fig:pareto-fronts}
\end{figure*}


We further explore the Pareto Front of performance and efficiency for all hybrid models considered herein. We relate the FLOPs of the token mixers to the language modeling performance for each model and hybridization setting (averaged over all benchmarks and both model sizes). Details for the FLOP calculation are presented in Appendix \ref{sec:flop_analysis_token_mixers}.
Figure \ref{fig:pareto-front-4k} shows this Pareto Front with the FLOPs calculated for the 1.3B parameter models and a sequence length of 4,096. The Pareto Front of optimality is shown with a dashed line. It is evident that the pure HGRN model, due to its vector-sized state, uses orders of magnitude fewer FLOPs than all other models. The full transformer model is on the opposite end of the Pareto Front, representing the highest performance and lowest efficiency. The middle of the Pareto Front is occupied by the HGRN 24:1 model and the HGRN2 6:1 model. 

Figure \ref{fig:pareto-front-32k} shows the same models with the FLOPs calculated for a sequence length of 32,768. This pushes the efficiency of models with attention layers lower. The two ends of the spectrum--the pure HGRN and the pure transformer--remain, but the lower-performance half of the spectrum is now also occupied by the pure RetNet, pure HGRN2, and pure GatedDeltaNet (in increasing order of performance).

It is important to note that our efficiency metric--FLOPs in the forward pass--does not directly translate into throughput or latency on conventional hardware. Specifically, the pure HGRN model is the most efficient model in our analysis but this efficiency may not be reflected in throughput on modern GPUs. Although the HGRN requires fewer operations in total, it still requires a comparable number of memory accesses and its element-wise vector operations may under utilize the GPU in comparison to matrix operations in other token mixers. However, the HGRN's reduced computational footprint may be accelerated on smaller GPUs or neuromorphic hardware \cite{zhu2024scalablematmulfreelanguagemodeling,abreu2025neuromorphic}.

\section{Forward-Pass FLOP Analysis of Token Mixers}
\label{sec:flop_analysis_token_mixers}

This section derives closed‑form expressions for the floating‑point operations (FLOPs) required by the different recurrent and linear‑attention token mixers that are presented in this paper, and contrasts them with classical softmax self‑attention.  All counts are \textbf{per layer, forward pass only}, aggregated over all heads.  Projection layers, (un-)embedding, residual additions, normalization, and other feed‑forward blocks are \emph{not} included so that every token mixer is compared on equal footing.

\paragraph{Notation and assumptions}
\begin{itemize}
  \item $L$ – causal sequence length (tokens seen by the layer).
  \item $H$ - number of heads in the model (this applies to all trained models except the HGRN-1 model where $H=1$ by design).
  \item $d$ – width of a single head. Vectors within a single head live in $\mathbb{R}^{d}$. For a model dimension $d_\text{model}$ with $H$ heads we have $d = d_\text{model}/H$.
  \item One FLOP is a single floating‑point multiply \emph{or} add.
  \item Only the token-mixer is costed.  Projections, MLPs, norms, residuals are identical across models and therefore omitted.
  \item All counts below are \emph{per token, per layer, summed over all heads}.  Multiply by $L$ for a sequence, and by the number of layers to estimate a full network.
\end{itemize}

\begin{table}[htbp]
  \centering
  \footnotesize
  \begin{tabular}{lcc}
    \toprule
    \textbf{Model} & \textbf{Total FLOPs / token} & \textbf{Complexity in $L$} \\
    \midrule
    Softmax self-attention            & $\displaystyle 2L d_{\mathrm{model}}$                          & $\Theta(L^2 d_{\mathrm{model}})$ \\
    \midrule
    HGRN-1 (vector, $H\!=\!1$)        & $\displaystyle 5 d_{\mathrm{model}}$                            & $\Theta(L d_{\mathrm{model}})$ \\
    Hawk (RG-LRU, vector)              & $\displaystyle 5 d_{\mathrm{model}}$                            & $\Theta(L d_{\mathrm{model}})$ \\
    \midrule
    RetNet / Lightning                 & $\displaystyle 5\,\frac{d_{\mathrm{model}}^{2}}{H}$            & $\Theta\bigl(L d_{\mathrm{model}}^{2}/H\bigr)$ \\
    Mamba-2                            & $\displaystyle 5\,\frac{d_{\mathrm{model}}^{2}}{H}$            & $\Theta\bigl(L d_{\mathrm{model}}^{2}/H\bigr)$ \\
    GLA                                & $\displaystyle 7\,\frac{d_{\mathrm{model}}^{2}}{H}$            & $\Theta\bigl(L d_{\mathrm{model}}^{2}/H\bigr)$ \\
    RWKV-6                             & $\displaystyle 7\,\frac{d_{\mathrm{model}}^{2}}{H}$            & $\Theta\bigl(L d_{\mathrm{model}}^{2}/H\bigr)$ \\
    HGRN-2 / MetaLA                    & $\displaystyle 7\,\frac{d_{\mathrm{model}}^{2}}{H}$            & $\Theta\bigl(L d_{\mathrm{model}}^{2}/H\bigr)$ \\
    DeltaNet                           & $\displaystyle 8\,\frac{d_{\mathrm{model}}^{2}}{H}$            & $\Theta\bigl(L d_{\mathrm{model}}^{2}/H\bigr)$ \\
    Gated DeltaNet                     & $\displaystyle 8\,\frac{d_{\mathrm{model}}^{2}}{H}$            & $\Theta\bigl(L d_{\mathrm{model}}^{2}/H\bigr)$ \\
    \bottomrule
  \end{tabular}
  \caption{Per-token forward FLOP counts \emph{summed over all heads}.  Multiply by $L$ to obtain the per-layer sequence cost.  For HGRN-1 the mixer is single-head by design ($H=1$); all other models allow arbitrary $H$.  Softmax attention is quadratic in $L$, whereas the linear/recurrent mixers are linear.}
  \label{tab:flops_summary_all_heads}
\end{table}

\paragraph{Baseline: softmax self-attention}
The dot-product matrix $QK^{\top}$ is strictly lower-triangular under causal masking.  Forming and applying it costs $(H\,L(L+1)d_{\mathrm{head}})$ multiplies plus the same number of adds.  Using $d_{\mathrm{model}} = H d_{\mathrm{head}}$ gives
\begin{equation}
  \mathrm{FLOPs}_{\text{softmax}} \;\approx\; 2L^2 d_{\mathrm{model}}.
  \label{eq:softmax_total_heads}
\end{equation}
Importantly the head count cancels, so the quadratic term depends only on the model width.

\paragraph{Vector-state mixers (classical / gated RNNs)}
A single hidden vector $h_t\!\in\!\mathbb{R}^{d_{\mathrm{head}}}$ is maintained \emph{per head}.  Each head therefore needs $5d_{\mathrm{head}}$ FLOPs per token. Aggregating over $H$ heads:
\begin{equation}
  \mathrm{FLOPs}_{\text{vector}} = 5H d_{\mathrm{head}} = 5 d_{\mathrm{model}}.
\end{equation}
For HGRN-1 and Hawk, the reference implementation is single-head; we therefore take $H=1$ and $d_{\mathrm{head}} = d_{\mathrm{model}}$.

\paragraph{Matrix-state mixers (outer-product family)}
Each head carries a dense state $S_t \in \mathbb{R}^{d_{\mathrm{head}}\times d_{\mathrm{head}}}$.
If an update costs $k\,d_{\mathrm{head}}^{2}$ FLOPs per head (where $k$ is a constant that varies between different models), the layer total is:
\begin{equation}
  \mathrm{FLOPs}_{\text{matrix}} = k H d_{\mathrm{head}}^{2} \;=\; k\,\frac{d_{\mathrm{model}}^{2}}{H}.
  \label{eq:matrix_family_general}
\end{equation}
Hence adding more heads \emph{reduces} the cost of these mixers at fixed $d_{\mathrm{model}}$.
Each ``pass'' over the $d_{\mathrm{head}} \times d_{\mathrm{head}}$ matrix either multiplies all entries or adds a rank--1 outer product, each costing $d_{\mathrm{head}}^{2}$ FLOPs.  Summing the passes per token per head gives:
\begin{itemize}
  \item \textit{RetNet / Mamba--2}: five passes $\Rightarrow$ k=5.
  \item \textit{GLA, RWKV--6, HGRN--2}: two extra gating passes (seven total) $\Rightarrow$ k=7.
  \item \textit{DeltaNet (\& Gated)}: one further forget/restore pass (eight total) $\Rightarrow$ k=8.
\end{itemize}

\begin{figure}[htbp]
    \centering
    \includegraphics[width=0.8\linewidth]{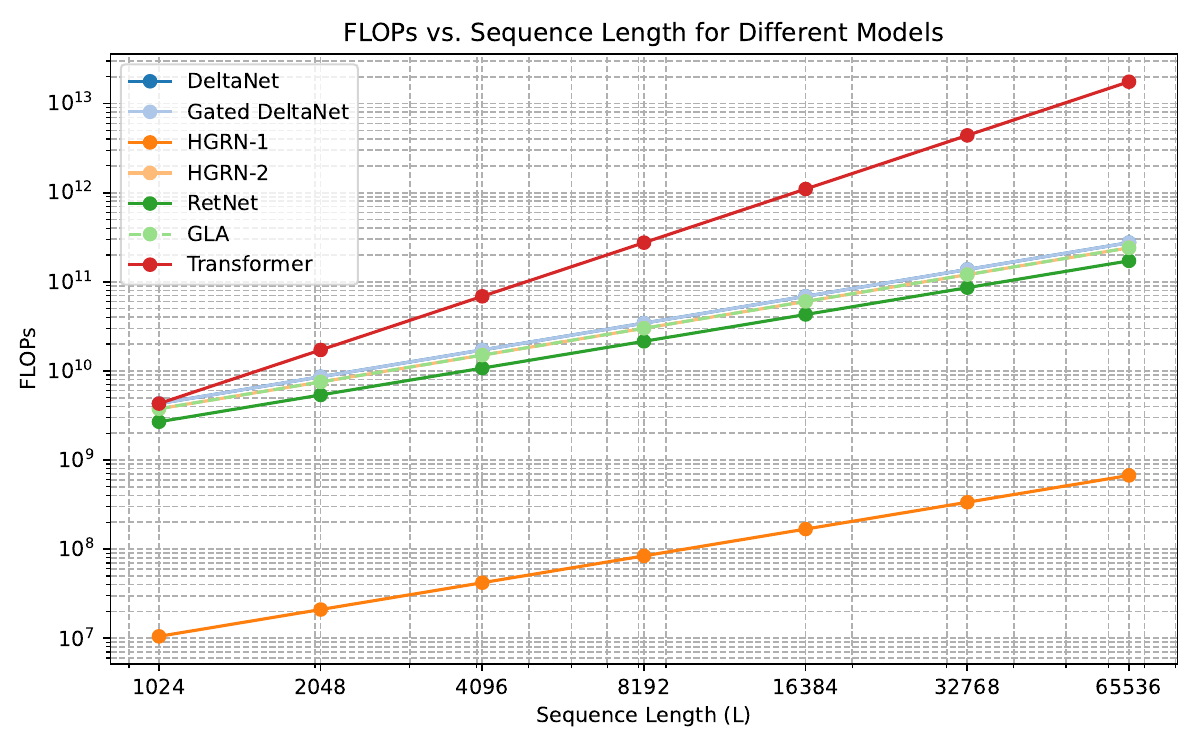}
    \caption{Relationship between the sequence length and the number of FLOPs required by different token mixers. Note that the HGRN-2 and GLA overlap, see analysis in the text.}
    \label{fig:flop-scaling}
\end{figure}

\paragraph{Summary}
Table \ref{tab:flops_summary_all_heads} shows a comparison between all models presented in Table 1. Once head aggregation is included the classical vector mixers (HGRN-1, Hawk) remain linear in $L$ and independent of $H$.  Outer-product and Delta-rule mixers grow as $d_{\mathrm{model}}^{2}/H$—wider models prefer more heads from a FLOP viewpoint—yet are still asymptotically cheaper than softmax self-attention for long sequences. Figure \ref{fig:flop-scaling} shows how the FLOPs scale with sequence lengths for different pure (non-hybrid) token mixers.

\section{RULER Results and Tables}
\label{sec:RULER_Results_and_Tables}


\begin{figure}[htbp]
    \centering
    \includegraphics[width=.7\linewidth]{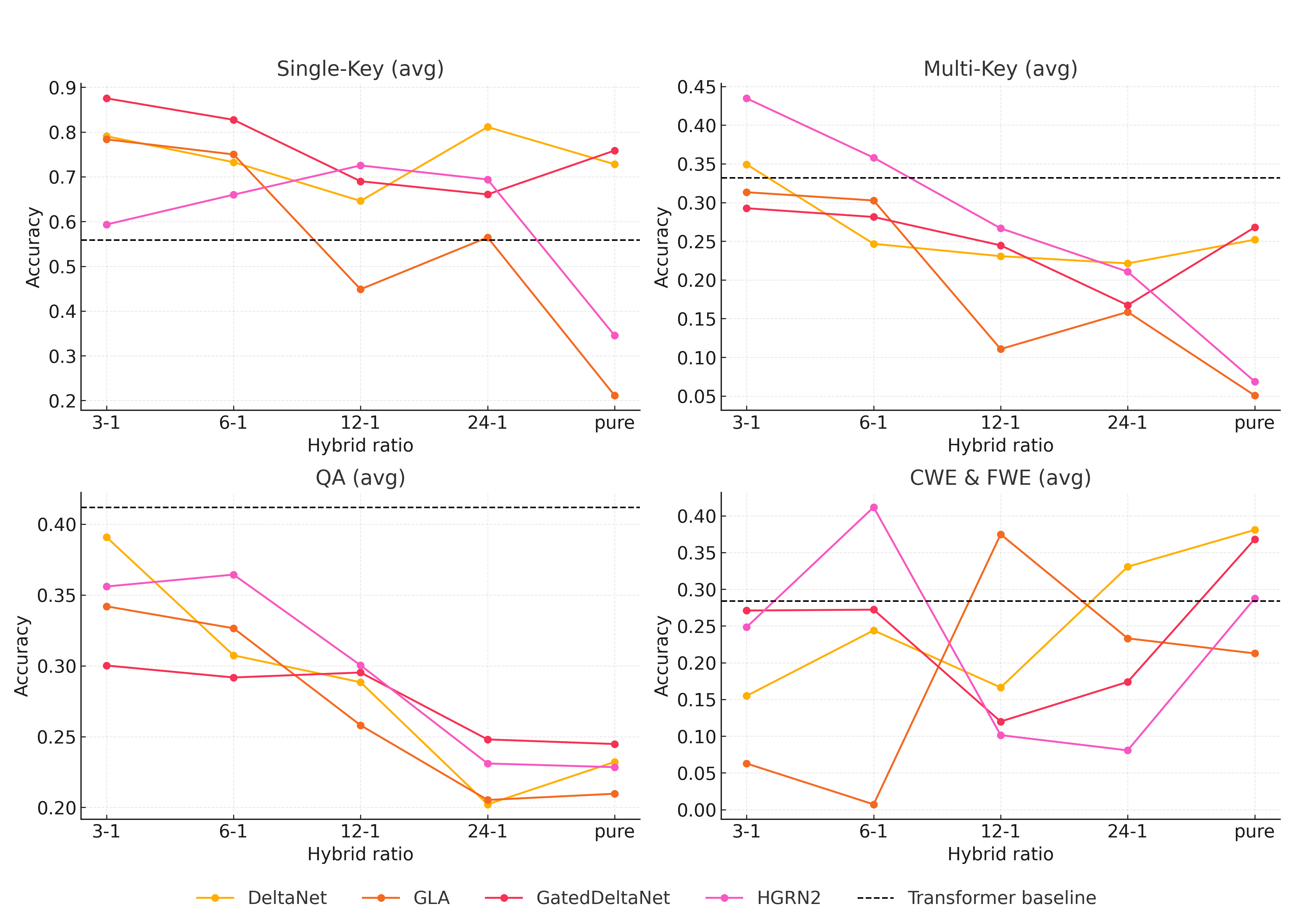}
    \caption{RULER sub task results based on ratio. RetNet and HGRN model families are omitted as their recall benchmark results were insignificant.}
    \label{fig:RULERsubtasks}
\end{figure}

\begin{table}[htbp]
\begin{center}
\footnotesize
\begin{tabular}{l|cc|cc|cc|cc}
\toprule
\textbf{Model} & \textbf{Multikey} & \textbf{Multiquery} & \textbf{Multivalue} & \textbf{Singlevalue} & \textbf{CWE+FWE} & \textbf{QA} & \textbf{VT} & \textbf{Avg}\\\midrule
DeltaNet-hybrid-3-1 &0.349 &0.471 &0.361 &0.791 &0.155 &0.391 &0.191 &0.426\\
GLA-hybrid-3-1 &0.313 &0.22 &0.196 &0.784 &0.063 &0.342 &0.234 &0.365\\
GatedDeltaNet-hybrid-3-1 &0.293 &0.542 &0.502 &0.875 &0.271 &0.300 &0.024 &0.440\\
HGRN2-hybrid-6-1 &0.358 &0.509 &0.507 &0.660 &0.412 &0.364 &0.014 &0.434\\
Transformers &0.332 &0.464 &0.472 &0.559 &0.284 &0.412 &0.066 &0.423\\
\bottomrule
\end{tabular}
\vspace{2mm}
\caption{Models from each family with the best average RULER results are shown here.}
\label{tab:RecallBestModelComparison}
\end{center}
\end{table}

We investigate the impact of the hybrid ratio further in Figure~\ref{fig:RULERsubtasks}. Single-Key, Multi-key, and QA sub tasks
are noticeably affected by a changing ratio, with higher concentrations of full attention performing better
than lower concentrations. Common Word Extraction (CWE) and Frequent Word Extraction (FWE) does
not seem to correlate with the hybrid ratio.

These trends indicate that, apart from pure recall tasks, where the goal is to remember some information
presented in the past of a long sequence, the hybrid ratio does not make much difference. Practitioners can
freely optimize the linear to full attention ratio with minimal effect on language modeling performance.

Table~\ref{tab:RecallBestModelComparison} shows a detailed view of RULER sub task results for the best performing ratios from each model family.

\section{Hybridization Strategy}
\label{sec:hybrid}

\begin{figure}[!h]
  \centering
  \includegraphics[width=.60\linewidth]{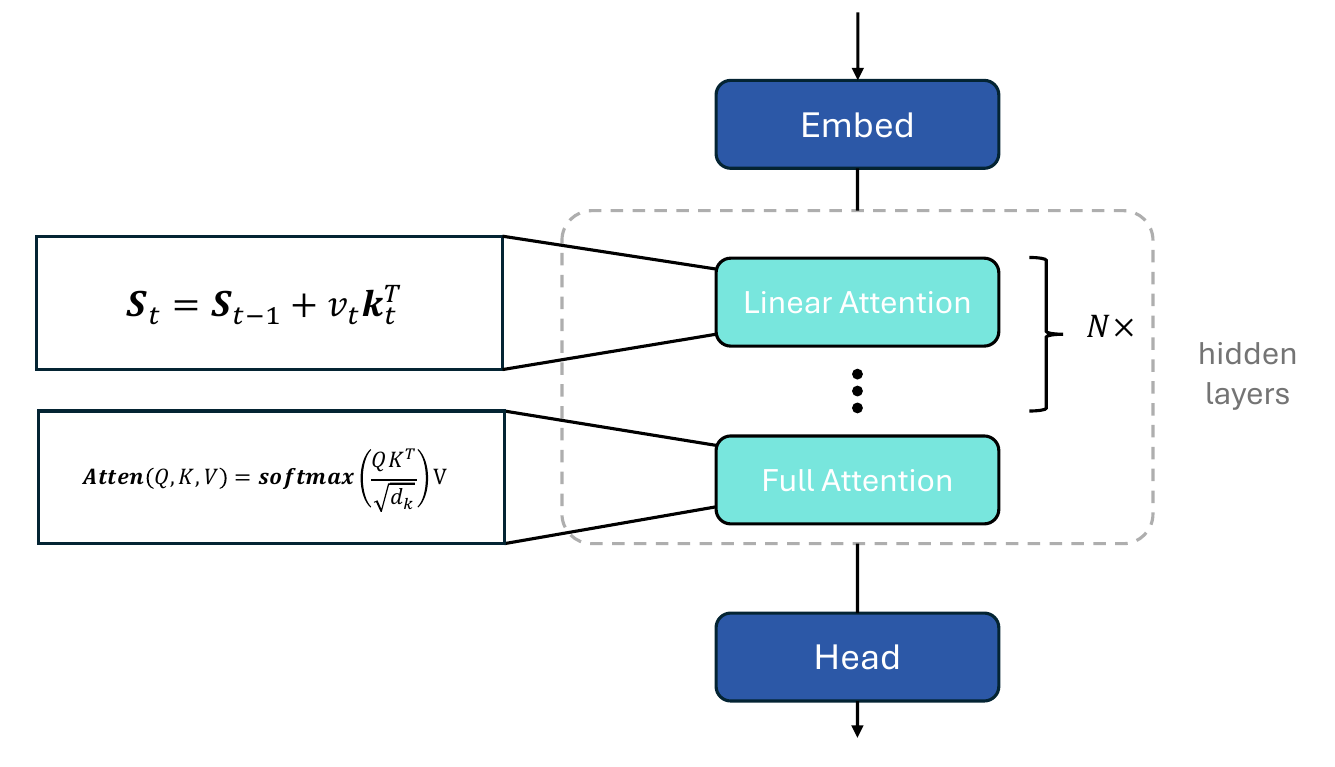}
  \vspace{-1mm}
  \caption{Hybrid architecture: an embedding layer, \(N\) repetitions of
  linear-attention followed by full-attention, and a projection head.
  Only the full-attention blocks grow the KV cache.}
  \label{fig:hybrid-arch}
\end{figure}

Figure~\ref{fig:hybrid-arch} sketches the hybrid stack used throughout our experiments.  
Each input sequence is first mapped to embeddings, after which the network applies a
repeating block of two conceptually different layers:

\begin{itemize}
    \item \textbf{Linear–attention layer.}  
    The layer carries a constant-size state \(\mathbf{S}_t\) that is updated by an outer-product write
    \(\mathbf{S}_t=\mathbf S_{t-1}+ \vv_t\vk_t^{\top}\)
    (or the model-specific variant from Table~\ref{tab:gen-comparison}).
    Because \(\mathbf{S}_t\) does not grow with the sequence, these layers add \(\mathcal O(Ld)\) or \(\mathcal O(Ld^2)\) compute and
    a negligible amount of cache at inference time.
    
    \item \textbf{Full-attention layer.}  
    A standard soft-max attention  
    \(\text{Atten}(Q,K,V)=\mathrm{softmax}\!\bigl(QK^{\top}/\sqrt{d_k}\bigr)V\)
    refreshes global token-to-token interaction and
    writes its keys and values to the KV cache.
\end{itemize}

We interleave the two kinds of layers in a fixed ratio
\(r{:}1\;\)(linear\,{:}\,full), repeating the composite block \(N\) times.
During training the stack is processed exactly like a Transformer;
during auto-regressive decoding only the full-attention layers enlarge the cache,
so the memory footprint is reduced by roughly a factor of \(r\).
In Section \ref{sec:emprical} we sweep \(r\in\{24,12,6,3\}\) (plus the pure cases) to quantify
how this trade-off affects language-model loss and recall capability.

\section{Ratio Comparison Tabular Results}
\label{sec:TableRatioResults}


\begin{table}[htbp]
\begin{center}
\footnotesize
\begin{tabular}{l|cc|cccc|c}
\toprule
\textbf{Ratio} & \textbf{DeltaNet} & \textbf{GatedDeltaNet} & \textbf{GLA} & \textbf{HGRN} & \textbf{HGRN2} & \textbf{RetNet} & \textbf{Average} \\
\midrule
24-1 &0.472 &0.468 &0.470 &\textbf{0.488} &0.488 &\textbf{0.472} &0.476 \\
12-1 &0.476 &0.478 &0.474 &0.482 &0.488 &\textbf{0.472} &0.478 \\
6-1 &\textbf{0.478} &0.481 &0.471 &0.486 &\textbf{0.494} &\textbf{0.472} &\textbf{0.480} \\
3-1 &0.475 &0.478 &\textbf{0.478} &0.483 &0.489 &0.468 &0.479 \\
pure &0.472 &\textbf{0.487} &0.459 &0.459 &0.486 &0.471 &0.472 \\
\bottomrule
\end{tabular}
\vspace{2mm}
\caption{Language modeling performance results from each linear attention variant and ratio are averaged across 1.3B and 340M parameter scales. Total average represents the aggregate score for each ratio. \textit{Pure} refers to a pure Linear Attention model.}
\label{tab:AggregateRatioLMComparison}
\end{center}
\end{table}

\begin{table}[htbp]
\begin{center}
\footnotesize
\begin{tabular}{l|cc|cc|c}
\toprule
\textbf{Ratio} & \textbf{DeltaNet} & \textbf{GatedDeltaNet} & \textbf{GLA} & \textbf{HGRN2} & \textbf{Average} \\
\midrule
24:1 &0.417 &0.341 &0.293 &0.303 &0.338 \\
12:1 &0.352 &0.352 &0.288 &0.368 &0.340 \\
6:1 &0.371 &0.425 &0.330 &\textbf{0.434} &0.390 \\
3:1 &\textbf{0.426} &\textbf{0.436} &\textbf{0.332} &0.395 &\textbf{0.397} \\
Pure &0.320 &0.356 &0.144 &0.203 &0.256 \\
\bottomrule
\end{tabular}
\vspace{2mm}
\caption{Recall performance results from each linear attention variant and ratio are averaged. Total average represents the aggregate score for each ratio.}
\label{tab:AggregateRatioRecallComparison}
\end{center}
\end{table}

Table~\ref{tab:AggregateRatioLMComparison} and \ref{tab:AggregateRatioRecallComparison} are tabular versions of the results depicted in Figure~\ref{fig:trend-graph}.

\end{document}